\documentclass{article}
\usepackage{spconf,amsmath,amssymb,graphicx}
\usepackage{times}
\usepackage{booktabs}
\usepackage{multirow}
\usepackage{algorithm}
\usepackage{algorithmic}
\usepackage{cite}
\title{Deep-Prior-Guided Superpixel Segmentation of Choroidal Nevi\\ in High-Resolution Fundus Images}

\makeatletter
\gdef\@name{{\em Mohammadmahdi Eshragh$^{1}$, Emad A.\ Mohammed$^{2}$, Behrouz Far$^{1}$, Ezekiel Weis$^{3}$,}\\
            {\em Carol L.\ Shields$^{4}$, Sandor R.\ Ferenczy$^{4}$, Trafford Crump$^{5}$%
            \thanks{Corresponding author: mohammadmahdi.eshrag@ucalgary.ca}}\\}
\makeatother

\address{\normalsize $^{1}$Dept.\ of Electrical \& Software Engineering and $^{5}$Dept.\ of Surgery, University of Calgary, Canada\\
         \normalsize $^{2}$Dept.\ of Computer Science and Physics, Wilfrid Laurier University, Waterloo, Canada\\
         \normalsize $^{3}$Dept.\ of Ophthalmology \& Visual Sciences, University of Alberta, Edmonton, Canada\\
         \normalsize $^{4}$Wills Eye Hospital, Philadelphia, PA, USA}

\begin{document}
\ninept
\maketitle

\begin{abstract}
Choroidal nevi (CN) are common benign fundus lesions that require longitudinal monitoring because a small fraction transform into melanoma. Accurate delineation at native resolution is challenging: CN typically occupy $<2\%$ of the image, boundaries are indistinct, and expert annotations are limited. We propose a hybrid pipeline in which an unmodified UNet trained on $128\times128$ images provides a coarse location prior, while a training-free superpixel method (SLIC) recovers the boundary at the full input resolution. The prior further drives automatic per-image selection of the superpixel count via an image-to-lesion ratio, eliminating manual tuning. On 25 held-out $1024\times1024$ images, the proposed method achieved a mean Dice of 90.9\% (IoU 87.9\%), outperforming nnU-Net (83.2\%), Swin UNet (73.0\%), and Attention UNet (60.6\%) trained at the same resolution, with all paired differences significant after Holm correction. On 90 external images from a second centre, the advantage persisted (76.8\% vs.\ 69.3\% for nnU-Net). Ablation studies identify the automated superpixel count as the key component. Inference needs no GPU, requiring 1.6\,s per $1024\times1024$ image and 15\,s at $3900\times3900$ on a laptop.
\end{abstract}

\begin{keywords}
Choroidal nevus, fundus photography, lesion segmentation, superpixels, data-efficient deep learning
\end{keywords}

\section{Introduction}
\label{sec:intro}

Choroidal nevi (CN) are pigmented lesions of the posterior segment found in a substantial share of adults. Although benign, a small fraction transform into uveal melanoma, whose incidence is about six cases per million person-years~\cite{gomez_health_2024}, and delayed detection of melanoma carries a risk of vision loss, metastasis, and death~\cite{cheung_distinguishing_2012}. Automated delineation on routine colour fundus photographs would support consistent measurement of lesion size and growth over time~\cite{crump_artificial_2025}.

Segmenting CN is harder than segmenting the optic disc or vessels. Lesions blend gradually into the surrounding choroid, contrast is low, and even trained annotators place borders inconsistently~\cite{eshragh_enhancing_2024}. Annotations require ocular oncologists, and no public CN segmentation dataset exists~\cite{crump_artificial_2025}. Deep learning (DL) further adds a resolution problem: CN occupy on average less than 2\% of the image, so at native resolution ($\approx4000\times4000$ pixels) lesion signal is swamped by background structure; UNet-family models often lose accuracy as the input size grows while training time and energy rise steeply~\cite{eshragh_enhancing_2024}. Down-sampling improves learning but discards boundary detail needed for clinical measurement, and prior CN work therefore operated at reduced resolution~\cite{eshragh_enhancing_2024,biglarbeiki_enhancing_2024}. Self-configuring nnU-Net remains a strong general baseline~\cite{isensee_nnunet_2021} but relies on full-resolution training and GPU ensembling, while promptable foundation models such as SAM and MedSAM are least accurate on small, low-contrast targets with indistinct borders and require prompts and large backbones at inference~\cite{kirillov_segment_2023,ma_segment_2024,mazurowski_segment_2023}.

Classical superpixel methods sit at the opposite end of this trade-off. SLIC~\cite{achanta_slic_2012} clusters pixels by colour and spatial proximity, requires no training data, and scales to any resolution, but its parameters, especially the number of segments, must be tuned per image, and it has no notion of which segment is clinically relevant. Superpixels and CRFs have long refined coarse CNN outputs or pooled CNN features~\cite{chen_deeplab_2018,jampani_ssn_2018}, but with the network trained or run at the target resolution; here it never sees more than $128\times128$ pixels, and its output sets the superpixel scale per image (Eq.~\eqref{eq:K}), which the ablation shows is what matters.

We combine the two so that each operates where it is strongest: a lightweight, unmodified UNet localises the lesion from a $128\times128$ input, and SLIC delineates its boundary at the input resolution. The learned mask serves only as a prior that both fixes the superpixel count through an image-to-lesion ratio and selects the superpixel to keep. We contribute (i) a training-free, CPU-only delineation stage driven by a low-resolution learned prior; (ii) a paired per-image comparison against nnU-Net, Swin UNet, and Attention UNet; (iii) external validation on 90 images from a second centre with a covariate analysis of domain shift; and (iv) an ablation showing that per-image adaptation of the superpixel count is the component that matters.

\section{Method}
\label{sec:method}

\subsection{Location prior}
The guide network $f$ is a standard UNet~\cite{ronneberger_u-net_2015} with no architectural modifications. It is trained on $128\times128$ RGB images scaled to $[0,1]$ using flips, rotations, random crops, additive noise, colour jitter, and elastic deformations; the learning rate ($10^{-4}$), batch size, and loss (Dice vs.\ focal) are selected by grid search with five-fold cross-validation on the training portion. The sigmoid output is thresholded at 0.5 to obtain a binary mask $M$, which is then resized to the full input size by nearest-neighbour interpolation to produce $M_F$. Operating at $128\times128$ keeps training and inference lightweight and CPU-deployable, and the network is tasked only with coarse localisation rather than pixel-accurate boundary tracing.

\subsection{Image-to-lesion ratio}
Let $A_{\text{lesion}}$ denote the area of the largest connected component of $M_F$, let $A_{\text{total}}$ be the image area, and define the image-to-lesion ratio as $R=A_{\text{total}}/A_{\text{lesion}}$. We then set the number of superpixels to
\begin{equation}
K=\mathrm{Coef}\cdot\mathrm{round}\!\left(cR/\mathrm{Coef}\right),
\label{eq:K}
\end{equation}
where $\mathrm{Coef}=A_{\text{total}}/A_{\text{largest training lesion}}$ ($=5$ here) is the smallest ratio observed in training and thus acts as both the minimum value and the quantisation step of $K$, and $c=0.85$ is a calibration factor chosen by grid search on the validation set jointly with the SLIC compactness (11) and Gaussian smoothing $\sigma$ (2). These constants are fixed before inference, so no dataset-level statistics are required at test time; $K$ is unbounded above and ranged from 15 to 930 (median 135) on the internal test set. If $M_F$ is empty, $R$ is undefined, so we set $K\leftarrow K_{\text{default}}=700$ (the ratio of the smallest training lesion); because no superpixel can overlap an empty prior, the pipeline returns an empty mask and flags the image for manual review as a deterministic deferral rule.

\subsection{Superpixel generation and selection}
SLIC partitions the full-resolution image $I$ (in CIELAB space) into $S=\mathrm{SLIC}(I,K,\text{compactness},\sigma)$ superpixels~\cite{achanta_slic_2012}. Since SLIC is purely appearance-driven and has no semantic notion of the lesion, we score each superpixel $s_i$ by its agreement with the prior,
\begin{equation}
R_i=\frac{1}{|s_i|}\sum_{p\in s_i}M_F(p),\qquad s_{\text{best}}=\arg\max_i R_i,
\label{eq:Ri}
\end{equation}
and return $I_{\text{seg}}(p)=\mathbf{1}[p\in s_{\text{best}}]$, followed by morphological closing with a $10\times10$ kernel. Algorithm~\ref{alg:hybrid} summarises the complete procedure; it uses only scikit-image and OpenCV functions, and an independent re-implementation from this description reproduced all 25 internal test masks exactly. Thus the UNet operates where the signal-to-noise ratio is highest, SLIC where fine structure is preserved, and Eq.~\eqref{eq:K} matches the superpixel scale to the lesion so that one superpixel can cover it.

\begin{algorithm}[t]
\caption{Hybrid segmentation of one fundus image}
\label{alg:hybrid}
\begin{algorithmic}[1]
\REQUIRE image $I$; guide network $f$; constants $\mathrm{Coef}$, $c$, $K_{\text{default}}$, compactness, $\sigma$
\STATE $M\leftarrow\mathbf{1}[f(\mathrm{resize}(I,128\times128))>0.5]$ \hfill\COMMENT{location prior}
\STATE $M_F\leftarrow\mathrm{resize}(M,\mathrm{shape}(I))$ \hfill\COMMENT{nearest neighbour}
\STATE \textbf{if} $M_F$ is empty: $K\leftarrow K_{\text{default}}$; \textbf{flag} for review; \textbf{return} empty mask
\STATE $A_{\text{lesion}}\leftarrow$ area of largest component of $M_F$; $R\leftarrow A_{\text{total}}/A_{\text{lesion}}$
\STATE $K\leftarrow\mathrm{Coef}\cdot\mathrm{round}(cR/\mathrm{Coef})$ \hfill\COMMENT{Eq.~\eqref{eq:K}}
\STATE $S\leftarrow\mathrm{SLIC}(I,K,\text{compactness},\sigma)$
\STATE $R_i\leftarrow\sum_{p\in s_i}M_F(p)\,/\,|s_i|$ for each $s_i\in S$ \hfill\COMMENT{Eq.~\eqref{eq:Ri}}
\STATE $s_{\text{best}}\leftarrow\arg\max_i R_i$
\STATE $I_{\text{seg}}\leftarrow\mathbf{1}[p\in s_{\text{best}}]$; morphological closing ($10\times10$)
\RETURN $I_{\text{seg}}$
\end{algorithmic}
\end{algorithm}

\section{Experiments}
\label{sec:experiments}

\subsection{Data}
We used 253 anonymised Alberta Ocular Brachytherapy Program RGB fundus images ($3900\times3900$) with expert binary lesion masks, and generated resized versions from $128\times128$ to $1024\times1024$. Images were split 90/10 into train/test (seed 101112; $n=25$ test), and training was further split 80/20 into train/validation. External testing used 90 Wills Eye Hospital images ($\approx4000\times4000$, resized to $1024\times1024$) annotated with the same protocol. Lesion size is reported only as a fraction of the imaged fundus area (mean $\pm$ SD): $1.64\pm2.03$\% internally and $4.97\pm8.52$\% externally.

\subsection{Baselines and protocol}
Baselines were Attention UNet~\cite{oktay_attention_2018}, Swin UNet~\cite{cao_swin-unet_2021}, and nnU-Net (2D configuration, default planner, fold~0 of its five-fold split, default test-time mirroring; trained for 102\,min, i.e., short of its default 1000-epoch schedule)~\cite{isensee_nnunet_2021}, each trained once on the same split at $1024\times1024$ (V100; UNet-variant learning rate $10^{-5}$; batch size and loss by grid search). GPU memory precluded training at $3900\times3900$, so all accuracy comparisons use $1024\times1024$ inputs, a size at which end-to-end networks are disadvantaged (Sec.~3.3); nnU-Net, which configures its own patching and normalisation, is therefore the most informative reference. To quantify resolution effects, we also trained a standard UNet at $128^2$, $256^2$, $512^2$, and $1024^2$. Dice/IoU are means of per-image scores.

For each baseline we tested paired per-image differences $\Delta\text{Dice}$ with two-sided Wilcoxon signed-rank tests (exact for $n=25$; normal approximation for $n=90$; zero differences excluded), Holm-corrected within each dataset. We report percentile-bootstrap 95\% CIs ($10^4$ resamples) and rank-biserial correlations $r_{rb}$, and verified conclusions with a sign-flip permutation test of the mean ($2\times10^5$ resamples).

\begin{figure*}[t]
\centering
\includegraphics[width=0.74\textwidth]{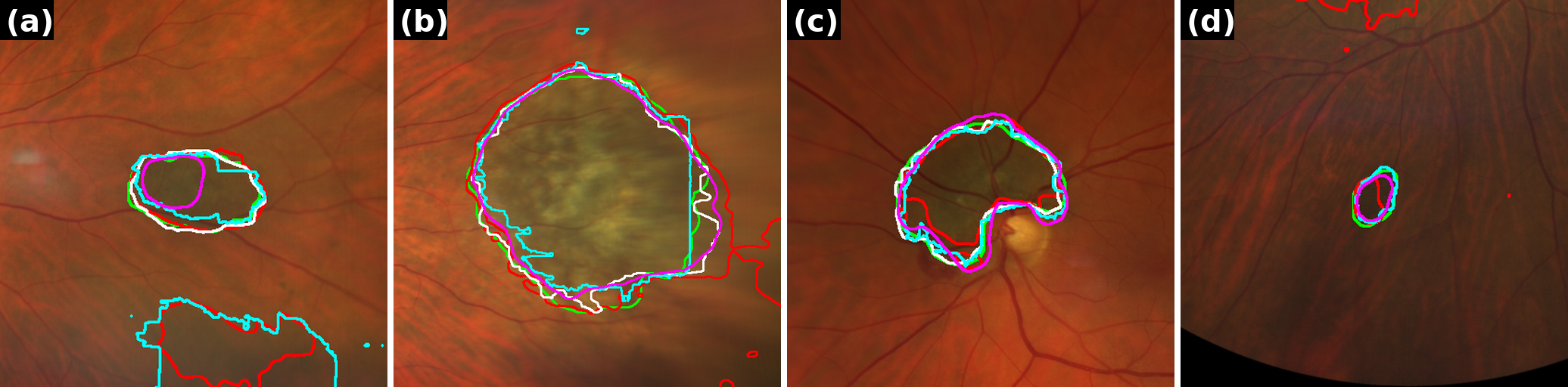}
\caption{Internal test examples ($1024\times1024$, lesion crops). Contours: ground truth (green), hybrid (white), nnU-Net (magenta), Swin (cyan), Attention (red). Dice (\%; hybrid/nnU-Net/Swin/Attention): (a) 93.4/47.1/21.8/25.6; (b) 93.6/92.0/90.7/82.2; (c) 94.7/90.9/91.5/83.7; (d) 0.0/43.6/7.0/1.6 (empty prior).}
\label{fig:qual}
\end{figure*}

\begin{table*}[t]
\centering
\footnotesize
\setlength{\tabcolsep}{3pt}
\caption{Per-image Dice and IoU at $1024\times1024$ on the internal ($n=25$) and external ($n=90$) sets; Dice as mean [bootstrap 95\% CI] and median, IoU as mean. Dice$\ge$80\%: images at or above 80\%. $\Delta$: hybrid minus model (Dice points). $p$: Wilcoxon signed-rank, Holm-corrected per dataset. $r_{rb}$: rank-biserial correlation. External ties (both masks empty) excluded before ranking.}
\label{tab:main}
\begin{tabular}{llccccccccc}
\toprule
 & & \multicolumn{2}{c}{Dice (\%)} & & & \multicolumn{2}{c}{$\Delta$Dice (points)} & & & \\
\cmidrule(lr){3-4} \cmidrule(lr){7-8}
 & Model & Mean [95\% CI] & Median & IoU (\%) & Dice$\ge$80\% & Median & Mean [95\% CI] & Wins/losses & $p$ (Holm) & $r_{rb}$ \\
\midrule
\multirow{4}{*}{\rotatebox{90}{Internal}}
 & Attention UNet & 60.57 [50.63, 69.22] & 65.89 & 47.05 & 6/25  & $+$31.45 & $+$30.36 [20.30, 40.42] & 23 / 2 & $4.5\!\times\!10^{-5}$ & 0.89 \\
 & Swin UNet      & 72.97 [63.06, 81.60] & 85.14 & 62.17 & 14/25 & $+$10.60 & $+$17.96 [8.89, 27.63]  & 21 / 4 & 0.0011 & 0.75 \\
 & nnU-Net        & 83.23 [77.73, 87.72] & 86.96 & 73.04 & 20/25 & $+$8.66  & $+$7.70 [0.93, 14.06]   & 21 / 4 & 0.0018 & 0.69 \\
 & \textbf{Hybrid (ours)} & \textbf{90.93} [81.27, 97.68] & \textbf{98.96} & \textbf{87.89} & \textbf{21/25} & -- & -- & -- & -- & -- \\
\midrule
\multirow{4}{*}{\rotatebox{90}{External}}
 & Attention UNet & 54.07 [48.19, 59.88] & 57.87 & 42.41 & 22/90 & $+$18.67 & $+$22.72 [16.35, 29.16] & 75 / 12 & $2.7\!\times\!10^{-9}$ & 0.76 \\
 & Swin UNet      & 59.28 [53.56, 64.93] & 63.19 & 47.27 & 26/90 & $+$14.83 & $+$17.51 [11.92, 22.95] & 73 / 13 & $8.9\!\times\!10^{-9}$ & 0.73 \\
 & nnU-Net        & 69.31 [63.14, 75.06] & 81.46 & 59.49 & 46/90 & $+$3.26  & $+$7.47 [2.81, 12.16]   & 72 / 15 & $4.8\!\times\!10^{-7}$ & 0.62 \\
 & \textbf{Hybrid (ours)} & \textbf{76.79} [69.94, 82.99] & \textbf{91.95} & \textbf{70.42} & \textbf{64/90} & -- & -- & -- & -- & -- \\
\bottomrule
\end{tabular}
\end{table*}

\subsection{Internal test set}
\begin{figure}[t]
\centering
\includegraphics[width=\columnwidth]{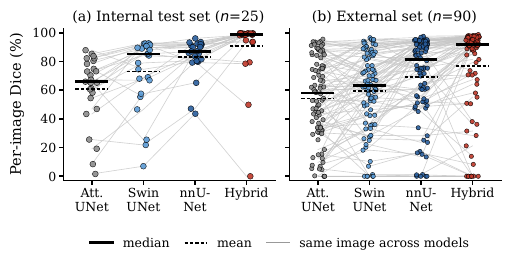}
\caption{Per-image Dice on (a) internal and (b) external sets (grey lines connect an image across models). Hybrid scores cluster near 99\% internally and above 90\% externally, with 15 external failures $<50$\% (7 with empty guide masks).}
\label{fig:perimage}
\end{figure}

The standard UNet was resolution-sensitive: 84.3\% Dice at $128\times128$ but 51.3\% at $1024\times1024$, while training time and energy rose about 40-fold from 3.2\,min and 15.8\,Wh. The hybrid instead uses the $128\times128$ UNet only for localisation and refines at the input resolution, reaching 90.93\% mean Dice (87.89\% IoU), 98.96\% median Dice, and the lowest Hausdorff distance (21.6\,px vs.\ 49.2/104.3/77.9\,px for nnU-Net/Swin/Attention; Table~\ref{tab:main}, Fig.~\ref{fig:perimage}a). It was best on 21/25 images.

The upsampled $128\times128$ guide mask alone (76.5\%, Table~\ref{tab:ablation}, E) already exceeds Swin UNet and Attention UNet at $1024\times1024$, which quantifies the price those architectures pay for training at this size; the comparison with nnU-Net is the demanding one, and the hybrid's median margin over it is $+8.7$ Dice points internally and $+3.3$ externally.

Permutation tests supported all paired comparisons ($p=0.033$, $6.1\times10^{-4}$, $1.5\times10^{-5}$ vs.\ nnU-Net, Swin, Attention); with $n=25$ and one training run per model, the margin over nnU-Net (95\% CI 0.9--14.1 points) is indicative, and the external set (Sec.~3.4) is the confirmatory test. The two outliers are an empty guide mask (deferred; Fig.~\ref{fig:qual}d, all baselines also failed) and a low-contrast lesion on which Attention UNet reached 87.8\%. When the prior is correct, the selected superpixel closely follows the boundary (21 images with Dice $\ge93.4$\%; Fig.~\ref{fig:qual}a).

\subsection{External validation}
All methods degraded externally, with similar drops for the three strongest models (hybrid $-14.1$, nnU-Net $-13.9$, Swin $-13.7$ Dice points). Ranking was unchanged and the hybrid retained the same margin over nnU-Net ($+7.70$ internal vs.\ $+7.47$ external; Table~\ref{tab:main}, Fig.~\ref{fig:perimage}b). The pipeline was best on 69/90 images, and all paired comparisons remained significant after Holm correction ($p\le4.8\times10^{-7}$), so the larger sample confirms the internal result.

Within the fundus area, brightness/contrast/colour cast/sharpness did not differ between datasets (SMD$<0.2$; Mann--Whitney $p\in[0.28,0.73]$), but lesions were larger externally (4.97 vs.\ 1.64\% area; $p=0.0016$; SMD $+0.54$). Failures (Dice $<0.5$, $n=15$) had larger lesions (median 8.4\% vs.\ 2.0\%; $p=0.0012$), while lesion size and Dice were otherwise uncorrelated ($\rho=0.04$). This matches Eq.~\eqref{eq:Ri}: very large lesions may be split across multiple superpixels while only one is retained. Seven failures (and the analogous single internal case) were empty guide masks handled by deferral. Reinhard normalisation to training CIELAB statistics did not help (mean Dice $76.79\to75.59$\%).

\begin{table}[t]
\centering
\scriptsize
\setlength{\tabcolsep}{2.5pt}
\caption{Ablation on the internal test set ($n=25$; per-image Dice in \%). Variants are compared to the full model (A) by exact two-sided Wilcoxon tests, Holm-corrected over four tests. B$^*$ is an oracle over fixed $K$ values (not executable); C matched A on every image.}
\label{tab:ablation}
\begin{tabular}{llcccl}
\toprule
 & Variant & Mean Dice & Median & A better & $p$ (Holm) \\
\midrule
A & Full model (auto $K$, overlap rule) & 90.93 & 98.96 & -- & -- \\
B & Fixed $K=100$   & 69.12 & 85.94 & 21/25 & $4.0\!\times\!10^{-4}$ \\
B & Fixed $K=500$   & 45.70 & 39.42 & 23/25 & $6.9\!\times\!10^{-5}$ \\
B & Fixed $K=1000$  & 27.66 & 22.21 & 24/25 & $5.5\!\times\!10^{-5}$ \\
B$^*$ & Oracle best fixed $K$ per image & 80.37 & 90.34 & -- & -- \\
C & Automated $K$ + centroid rule & 90.93 & 98.96 & 0/25 & identical \\
E & Guide mask only (no SLIC) & 76.49 & 85.77 & 24/25 & $4.8\!\times\!10^{-7}$ \\
\bottomrule
\end{tabular}
\end{table}

\subsection{Ablation}
Fixed $K\in\{100,500,1000\}$ substantially reduced accuracy (mean Dice 69.12/45.70/27.66\%). Across 72 image--$K$ pairs, Dice dropped with scale mismatch $|\log_2 (K_{\text{auto}}/K_{\text{fixed}})|$ (Spearman $\rho=-0.86$, $p=2.8\times10^{-22}$), and even an oracle choosing the best fixed $K$ per image reached only 80.37\% mean Dice (Table~\ref{tab:ablation}); thus, per-image adaptation of $K$ is essential.

A centroid-based selection rule produced identical masks to the overlap rule on all 25 images because the automated $K$ typically concentrates most of the prior within one superpixel (median 74\%). The upsampled guide mask alone achieved 76.49\% Dice at $1024\times1024$, so SLIC refinement adds 14.4 Dice points (95\% CI [11.3, 18.1]; better on 24/25 images, tied on the empty-mask case).

\subsection{Computational cost}
Training the guide network required 3.16\,min/15.8\,Wh (V100), versus 102\,min/510\,Wh (nnU-Net), 134\,min/671\,Wh (Attention), and 161\,min/804\,Wh (Swin). On a laptop CPU (Ryzen~7 5700U; TensorFlow~2.21, scikit-image~0.26, OpenCV~5.0), inference took 1.59\,s per $1024\times1024$ image (0.56\,s UNet, 1.00\,s SLIC) and 15.07\,s per $3900\times3900$ image (SLIC 14.16\,s; images upsampled from $1024^2$ for timing). Runtime was dominated by SLIC and scaled with pixel count ($14.5\times$ pixels, $14.1\times$ time), with little dependence on $K$ (Pearson $r=-0.09$). nnU-Net on CPU (PyTorch~2.14, test-time mirroring) required 16.98\,s per $1024\times1024$ image plus 23\,s start-up, so the proposed pipeline is faster even at native resolution and its 1.6\,s latency at $1024\times1024$ is compatible with on-device deployment.

\section{Discussion and Conclusion}
\label{sec:conclusion}
A lightweight, unmodified guide network can be used purely to parameterise a classical, training-free refinement stage and still outperform strong learned baselines (including nnU-Net) on an internal test set and, with tighter confidence intervals, on 90 external images from a second centre.

Key limitations follow directly from the mechanism. The pipeline inherits guide-network false negatives: empty priors are detected deterministically and should trigger deferral for human review. Priors that misestimate lesion area can also induce a mismatched superpixel count $K$ and are not yet flagged, and lesions far larger than those seen in training may be split across multiple superpixels while only one is retained. Immediate extensions are (i) retaining multiple superpixels with $R_i$ above a threshold and (ii) adding a size-aware deferral gate based on predicted lesion area and prior--superpixel overlap.

Evaluation constraints include a small internal test set ($n=25$) with a single training run per model, single-annotator internal masks, and baselines trained only at $1024\times1024$, where end-to-end networks are disadvantaged, with nnU-Net trained on a single fold with a shortened schedule rather than as its default five-fold ensemble; accuracy at native resolution has not yet been quantified against the native masks, since no baseline could be trained at that size, although the pipeline already runs at $3900\times3900$ (Sec.~3.6). While the method is fully specified by Algorithm~\ref{alg:hybrid} and the constants above, the current implementation is part of the proprietary NeviSight project; we plan to explore a public reference release where permitted.

\section{Compliance with Ethical Standards}
This retrospective study used anonymised fundus images and followed the Declaration of Helsinki. Ethics approval was granted by the Health Research Ethics Board of Alberta -- Cancer Committee (HREBA.CC-17-0625) for both datasets; data are not publicly available due to privacy restrictions.

\section{Acknowledgments}
This work was supported by the New Frontiers in Research Fund -- Exploration grant. The authors declare no relevant financial or non-financial interests. An AI assistant (Claude, Anthropic) was used for language editing and to assist in drafting analysis scripts; all scientific decisions and conclusions are the authors' own, and the authors take responsibility for the final manuscript.

\bibliographystyle{IEEEbib}
\bibliography{refs}

\end{document}